\documentclass{article}

\usepackage{amsmath,amsfonts,bm}

\def\eqref#1{equation~\ref{#1}}

\def\1{\bm{1}}

\def\vv{{\bm{v}}}
\def\vw{{\bm{w}}}

\def\evv{{v}}
\def\evw{{w}}

\DeclareMathAlphabet{\mathsfit}{\encodingdefault}{\sfdefault}{m}{sl}
\SetMathAlphabet{\mathsfit}{bold}{\encodingdefault}{\sfdefault}{bx}{n}

\newcommand{\R}{\mathbb{R}}

\usepackage{graphicx}
\usepackage{subcaption}
\usepackage{tikz}
\usepackage{booktabs} 
\usepackage{algorithmic}
\usepackage{todonotes}
\usepackage{url}
\usepackage{hyperref}

 \usepackage[dblblindworkshop, final]{neurips_2025}
\usepackage[utf8]{inputenc} 
\usepackage[T1]{fontenc}    
\usepackage{hyperref}       
\usepackage{url}            
\usepackage{booktabs}       
\usepackage{amsfonts}       
\usepackage{nicefrac}       
\usepackage{microtype}      
\usepackage{xcolor}         

\title{Modelling the Doughnut of social and planetary boundaries with frugal machine learning}

\author{%
  Stefano Vrizzi
  \\
  UB School of Economics\\
  Universitat de Barcelona\\
  \texttt{vrizzi@ub.edu} \\
   \And
   Daniel W. O'Neill \\
   UB School of Economics \\
   Universitat de Barcelona\\
   \texttt{oneill@ub.edu} \\
}

\begin{document}

\maketitle

\begin{abstract}
  The `Doughnut' of social and planetary boundaries has emerged as a popular framework for assessing environmental and social sustainability. Here, we provide a proof-of-concept analysis that shows how machine learning (ML) methods can be applied to a simple macroeconomic model of the Doughnut. First, we show how ML methods can be used to find policy parameters that are consistent with `living within the Doughnut'. Second, we show how a reinforcement learning agent can identify the optimal trajectory towards desired policies in the parameter space. The approaches we test, which include a Random Forest Classifier and $Q$-learning, are frugal ML methods that are able to find policy parameter combinations that achieve both environmental and social sustainability. The next step is the application of these methods to a more complex ecological macroeconomic model.   
\end{abstract}

\section{Introduction}

\subsection{The Doughnut of social and planetary boundaries}\label{sec:intro_doughnut}

Economic growth, typically measured by rising GDP, remains the primary goal for most national governments. However, research shows that once basic human needs are met, further increases in GDP do not improve wellbeing \citep{collste2021}. Moreover, this pursuit of economic growth exacerbates environmental pressures, as GDP is tightly coupled to environmental indicators such as CO$_{2}$ emissions and material use \citep{haberl2020systematic}.

In response, a number of `post-growth' approaches have emerged, whose aim is to improve human wellbeing within environmental limits \citep{kallis2025}. One approach, which has been gaining traction with policymakers, is the `Doughnut' of social and planetary boundaries developed by economist Kate Raworth \citep{raworth2017doughnut, fanning2025doughnut}. The Doughnut visualises sustainability in terms of a doughnut-shaped space where resource use is high enough to meet people’s basic needs, but not so high it transgresses planetary boundaries.

An active area in post-growth research is the development of \emph{ecological macroeconomic models} \citep{hardt2017ecological}. These models evaluate the performance of different policies on a wide range of environmental and social indicators, such as those in the Doughnut. Within these models, the goal is not to maximise GDP, but instead to achieve the threshold values for each of the social variables, without exceeding the boundary values for the environmental variables \citep{o2018good}.  

\subsection{Applying machine learning to ecological macroeconomic models}

Machine learning (ML) can potentially be used to improve ecological macroeconomic models in several ways (e.g. design, computational efficiency, calibration, policy search). Here, we focus on two core aspects where ML can be used with a model of the Doughnut: 1) policy search, and 2) transition to a desired policy state. We show proof-of-concept examples for each aspect (see \nameref{results}, sections \ref{find_the_doughnut} and \ref{via_the_doughnut} respectively), using a toy model (see \nameref{methods} and Appendix, section \ref{supplementary_material_computational_model}). Our goal is to validate frugal ML tools on a computational model where the ground truth is known and observable (Fig. \ref{fig:ground_truth}). Ultimately, the aim is to scale up these ML tools to a more complex ecological macroeconomic model, such as COMPASS (the Comprehensive Model of Planetary And Social Sustainability) \citep{vogel2025compass}. COMPASS allows for policies to be assessed against the Doughnut's multiple environmental and social indicators.

\section{Methods}\label{methods}

\begin{figure}[ht!]
    \centering
    \includegraphics[width=.45\linewidth,
    trim={0.0cm, 0.3cm, 0.0cm, 0.25cm}, clip
    ]{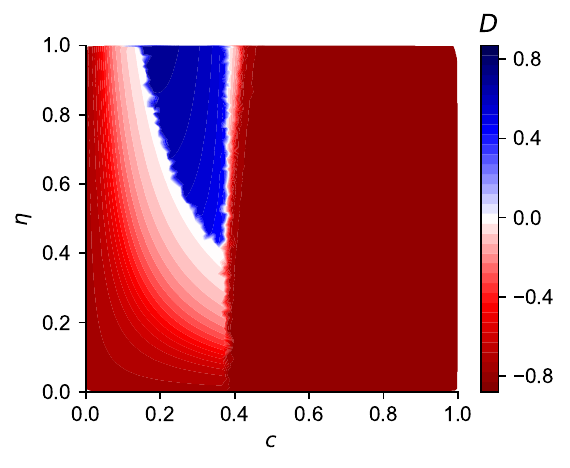}
    \caption{\textbf{
    `Doughnut' score of a computational toy model with respect to the two model parameters of interest: consumption \bm{$c$} and efficiency \bm{$\eta$}.
    }
    We consider a simple model (section \ref{supplementary_material_computational_model} in Appendix) and compute an aggregate performance measure, the Doughnut score $D$ (eq. \ref{eq:doughnut}). We consider a toy model because the ground truth of its Doughnut score's parameter-dependence is easily observable in two dimensions, unlike for complex economic models. The blue area represents the `Doughnut', i.e. the desired model output, which identifies the range of the corresponding desired model input parameters.}
    \label{fig:ground_truth}
\end{figure}

While true ecological macroeconomic models such as COMPASS contain many variables \citep{vogel2025compass}, we devise a simple consumer-resource model with one environmental indicator and one social indicator (section \ref{supplementary_material_computational_model} in Appendix) to explore our idea. Here, we optimise only two model parameters: 
consumption $c$ and efficiency $\eta$, which reflect two important policy levers for achieving sustainability. We define a vector $\vv = [\evv_1, ..., \evv_N]$ of components $v_n \in \R$ describing $N$ model performance scores, defined as the $N$ sustainability and socio-economic indicators of choice averaged over time (section \ref{supplementary_material_indicators} in Appendix). In our case, for simplicity, we study one environmental and one socio-economic performance score, i.e. $N=2$. We summarise aggregate performance by combining individual scores into a single `Doughnut' score (Fig. \ref{fig:ground_truth}):
\begin{equation}\label{eq:doughnut}
    D = \vw^T \vv + P(\vv) \vw^T \text{ReLU}(\vv),   
\end{equation}
where $\vw \in [0, 1]^N$ is a weight vector satisfying $\sum_{n=1}^N \evw_n = 1$ that scales the relative importance of individual performance scores, while $P(\vv) = \prod_{n=1}^N H(\evv_n) \, -1$ is a binary term based on the Heaviside step function $H$. We classify the model output into two cases: either all objectives are met ($\forall n,\; v_n > 0 \;\Rightarrow\; P(\vv) = 0$) or not ($\exists n: v_n \leq 0 \;\Rightarrow\; P(\vv) = 1$). In the first case, we say that the model dynamics `fall \textit{within} the Doughnut' (class `+'); $D>0$ and it scales with the magnitude of all indicators. In the second case, model dynamics `fall \textit{outside} the Doughnut' (class `-'), and only negative indicators contribute to $D$, hence $D\leq0$. 

\section{Results}\label{results}

\subsection{Locating the Doughnut: socio-economic and environmental policy search}\label{find_the_doughnut}

\begin{figure}[htp!]
    \centering
    \includegraphics[width=.8\linewidth,
    trim={0.0cm, 0.5cm, 0.0cm, 0.2cm}, clip
    ]{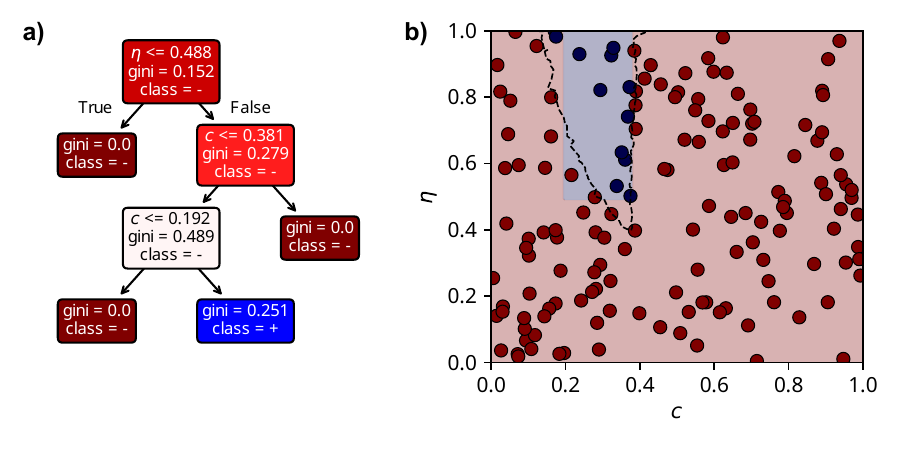}
    \caption{\textbf{Output from a Random Forest Classifier trained to find model input parameter values producing desired model ouputs (i.e. within the Doughnut).} Colour coding: red indicates the `-' class (i.e. model output falling outside the Doughnut),  blue indicates the `+' class (i.e. within the Doughnut). \textbf{a)} Decision path of a single tree showing the requirements to reach the Doughnut (blue box): high efficiency ($\eta > 0.488$), sustainable consumption ($c\leq 0.381$), and satisfying consumption needs ($c > 0.192$). \textbf{b)} Decision surface of the Random Forest Classifier represented as a background shade, where the colour indicates the predicted class. The circles are unseen test data, where the colour indicates the true label. The black dashed line circumscribes the area corresponding to the Doughnut ($D=0$ in Fig. \ref{fig:ground_truth}). 
    }
    \label{fig:RFC}
\end{figure}

We assume that we cannot access Fig. \ref{fig:ground_truth} (e.g. because of computational cost or high dimensionality). 
First of all, we want to find the model input parameter ranges that produce the outcomes of interest (the Doughnut), which may cover a small portion of the parameter space. A frugal ML tool in this case is a Random Forest Classifier (RFC), with model parameter inputs as features, and a binary output indicating whether the result is within or outside the Doughnut (see \nameref{methods}). The RFC correctly identifies the region of desired parameter values (Fig. \ref{fig:RFC}b), despite the high class imbalance and a max depth capped at 3 nodes. Moreover, it also provides interpretable decision paths (Fig. \ref{fig:RFC}a). Further details may be found in the Appendix (section \ref{sec:random_forest_classifier}).

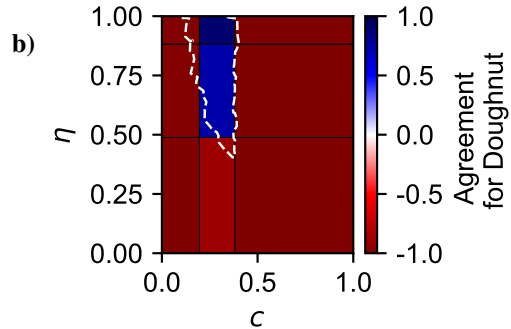
\begin{figure}[!htbp]
  \centering

   \begin{subfigure}[b]{0.48\textwidth}

    \centering
    \begin{tikzpicture}[remember picture,overlay]
      \node[anchor=north west] at (-.5,1.85) {\textbf{a)}}; 
    \end{tikzpicture}
    \begin{tabular}{ccc}
    \toprule
    $\eta$ & $c$ & Agreement \\
    \midrule
    0.882-1.000 & 0.196-0.381 & 0.903 \\
    0.488-0.882 & 0.196-0.381 & 0.738 \\
    0.000-0.488 & 0.196-0.381 & -0.887 \\
    0.488-0.882 & 0.000-0.196 & -0.986 \\
    0.882-1.000 & 0.000-0.196 & -0.990 \\
    0.488-0.882 & 0.381-1.000 & -0.994 \\
    0.000-0.488 & 0.000-0.196 & -1.000 \\
    \bottomrule
    \end{tabular}
    \vspace{.6cm}
    \label{tab:macro-params}
  \end{subfigure}
  \hfill

  \begin{subfigure}[b]{0.48\textwidth}
    \centering
    \begin{tikzpicture}[remember picture,overlay]
      \node[anchor=north west] at (-.35,4.05) {\textbf{b)}}; 
    \end{tikzpicture}    
    \includegraphics[width=.97\textwidth, trim={0.0cm, 0.5cm, 0.0cm, 0.2cm}, clip]{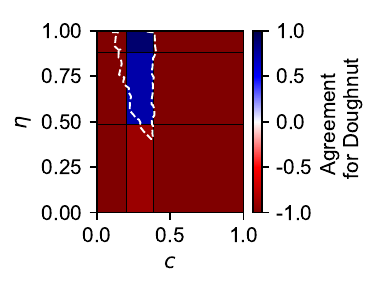}
    \label{fig:model-output}
  \end{subfigure}
  \caption{ 
  \textbf{Agreement scores on model output from model input parameter ranges.} We define `agreement' (see Appendix \ref{sec:agreement-computation}) as a score ranging from $-1$ (strong agreement for model output falling outside the Doughnut) to $+1$ (strong agreement for model output falling inside the Doughnut). Parameter ranges come from the RFC's trees (Appendix \ref{sec:agreement-computation}). \textbf{a)} Parameter ranges ranked by `agreement' in tabular form, yielding an interpretable and scalable solution (each model input parameter adds a column). \textbf{b)} Visual representation of the tabular form, to verify the agreement scores against the ground truth. The white dashed line indicates the Doughnut (D = 0 in Fig. \ref{fig:ground_truth}).
  }
  \label{fig:combined-panel}
\end{figure}

Since decision paths from individual trees can vary remarkably and be sensitive to noise, we propose a novel, frugal algorithm (Appendix~\ref{sec:agreement-computation}) that queries the entire forest, leveraging the RFC. The algorithm correctly identifies the desired model input parameter ranges (Fig. \ref{fig:combined-panel}b). The advantage with respect to RFC classification alone (Fig. \ref{fig:RFC}) is that parameter ranges can be ranked and presented in a tabular form (Fig. \ref{fig:combined-panel}a), which is both scalable and interpretable (one column, one parameter). The parameter ranges are defined by individual trees' thresholds, while the ranking is based on an `agreement' score computed from individual trees' classification, weighted by an accuracy measure on test data (Appendix~\ref{sec:agreement-computation}). This analysis can be complemented by feature importance ranking, which RFCs provide (Fig. \ref{fig:feature_importance} in Appendix).

\subsection{Reaching the Doughnut: policy transition path}\label{via_the_doughnut}

\begin{figure}[ht!]
    \centering
    \includegraphics[width=\linewidth,
    trim={0.0cm, 0.32cm, 0.0cm, 0.2cm}, clip
    ]{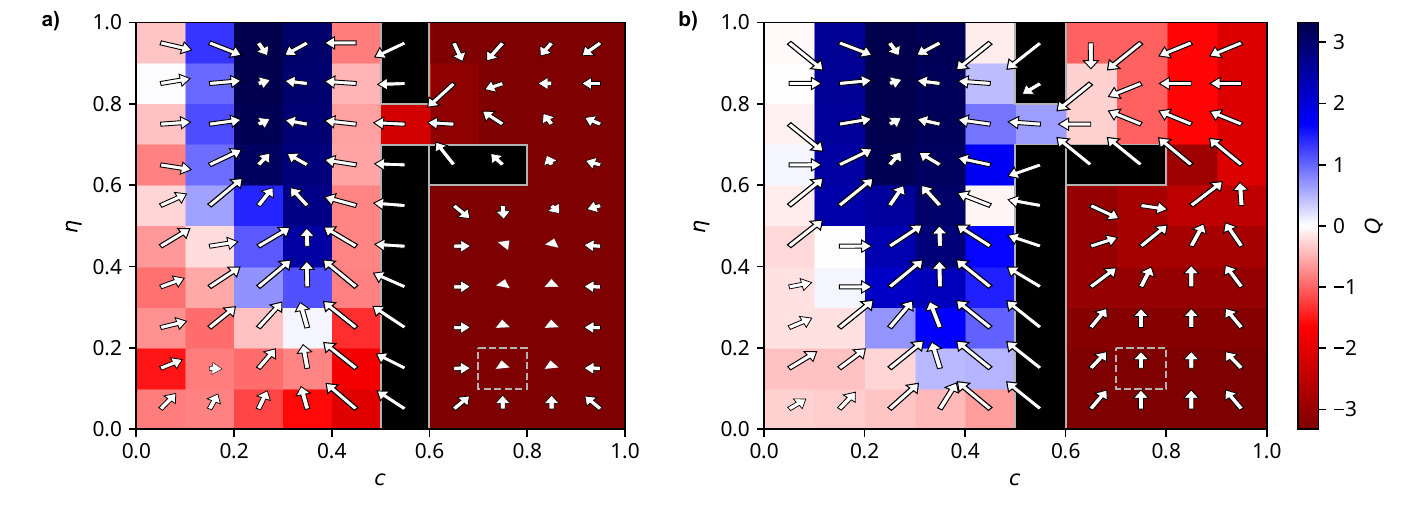}
    \caption{ \textbf{RL policy of the $Q$-learning agent after training.} 
    Each box is a state $s$ defined by discretising the space of model input parameters $c$ and $\eta$. The colour bar indicates the $Q$-value of action $a_{\text{stay}}$ after training. Arrows indicate the average action in each state.
    The agent successfully learns to avoid hypothetical barriers (i.e. strongly undesirable model outputs, shown in black), which were artificially set to further challenge the agent. The dashed grey line indicates the starting point of the agent. Panels \textbf{a} and \textbf{b} differ by discount factor ($\gamma=0.5$ and $\gamma=0.8$, respectively, see eq. \ref{eq:reward_q} in Appendix). The discount factor prioritises long-term over short-term rewards.}
    \label{fig:side_by_side_figures}
\end{figure}

In addition to locating the Doughnut, we can also study the optimal way to achieve the model output of interest from a given starting point. To do so, we train a reinforcement learning (RL) agent ($Q$-learning \citep{watkins1992q}) to maximise reward (eq. \ref{eq:reward_q}-\ref{eq:reward_q_2} in Appendix \ref{sec:q_learning}) based on the Doughnut score (eq. \ref{eq:doughnut}). The agent successfully reaches the desired model outputs, while avoiding the unwanted ones (Fig. \ref{fig:side_by_side_figures}a-b). Since the transition path here is long, with no gradient in local reward at the start, the discount factor $\gamma$ plays a key role (Fig. \ref{fig:side_by_side_figures}a-b).

\section{Discussion}\label{sec:discussion}

We briefly discuss two important points: 1) measuring performance, and 2) ML techniques. Concerning measuring performance, we use a scalar, aggregate variable ($D$, eq. \ref{eq:doughnut}). This aggregate variable is not fully consistent with the Doughnut, which is a `strong sustainability' framework \citep{o2018good}. Strong sustainability implies that higher performance on one indicator should not compensate for lower performance on another indicator \citep{neumayer2013}. While our approach ensures that all objectives are met, it is not strictly consistent with strong sustainability, and thus we continue to explore additional approaches.

Regarding ML techniques, each comes with its advantages and drawbacks \citep{molnar2018guide, scikit-learn}. For instance, our policy search (section \ref{find_the_doughnut}) can be optimised (e.g. \citealp{lamperti2018agent} and section \ref{sec:random_forest_classifier} in Appendix). 
In our policy-transition path search (section \ref{via_the_doughnut}), RL can be confronted with the curse of dimensionality in designing states and action spaces \citep{sutton2018reinforcement}, which affects the agent's exploration abilities. However, RL can also offer further advantages, such as hierachical RL \citep{barto2003recent} or co-operative agents (e.g. each agent optimises a different dimension of the Doughnut).

\section{Conclusions}\label{sec:conclusions}

Achieving sustainability is a complex and multi-dimensional challenge. Here, we have applied two standard ML techniques to highlight the potential of ML to improve ecological macroeconomic modelling. Specifically, we have presented two simple examples: 1) finding socio-economic policies (i.e. model input parameters) that are consistent with the Doughnut of social and planetary boundaries), and 2) identifying the optimal policy-transition path (i.e. traversing the model input parameter space). The next step is the application of these ML tools to COMPASS \citep{vogel2025compass}, a more complex ecological macroeconomic model that allows real policies to be assessed against the Doughnut's indicators.

\section*{Acknowledgments}
We would like to thank T. Domingos for his insightful comments on this paper. This research was funded by the European Union in the framework of the Horizon Europe
Research and Innovation Programme under grant agreement number 101137914 (MAPS: `Models,
Assessment, and Policies for Sustainability'). It received additional support from the AI4theSciences COFUND programme (Horizon 2020 grant agreement number 945304).


{
\small
\bibliography{references}
}

\newpage

\appendix

\section{Appendix}\label{Appendix}
\subsection{Computational model}\label{supplementary_material_computational_model}

We devise a simple dynamical system (eqs. \ref{eq:modelE} and \ref{eq:modelS}) based on two time-dependent variables, i.e. an environmental budget indicator $x_{\text{env}}$ and a social indicator $x_{\text{soc}}$: 

\begin{align}\label{eq:modelE}
\frac{dx_{\text{env}}}{dt} &= r \,  x_{\text{env}} \, (1 - x_{\text{env}}) \, H(x_{\text{env}}-x_{\text{env-crit}}) - \Tilde{c}, \\
\label{eq:modelS}
\frac{dx_{\text{soc}}}{dt} &= x_{\text{soc}} \, (1 - x_{\text{soc}}) \, \eta \, \Tilde{c} - \min\left(x_{\text{soc}}, c - \Tilde{c}\right),
\end{align}

with $x_{\text{env}}, x_{\text{soc}} \in [0, 1]$. We also define critical thresholds ($x_{\text{env crit}}$ and $x_{\text{soc crit}}$) for both indicators. The environmental threshold represents the critical environmental budget to preserve natural resources; the socio-economic threshold represents minimum socio-economic goals. The environmental threshold affects the model dynamics through the Heaviside step function $H$, reflecting the tipping point for the collapse of environmental resources. 
Model dynamics qualitatively depend on three parameters that are fixed for an individual simulation: consumption $c$ (which relates to the policy goal of sufficiency), efficiency $\eta$ (which relates to the policy goal of resource efficiency), and an environmental budget regeneration rate $r$ (e.g. the annual capacity of carbon sinks in climate change terms), here set to $r=1.5$. Moreover, we define $\Tilde{c} = \min(c, x_{\text{env}})$ to be actual consumption, as the environmental budget may be lower than consumption. In our analysis, we only focus on efficiency $\eta$ and consumption $c$, as they are the only two parameters in our simple model that are of interest from a socio-economic and environmental policy perspective.


\begin{figure}[htp!]
    \centering
    \includegraphics[width=.9\linewidth, 
    trim={0.0cm, 0.3cm, 0.0cm, 0.2cm}, clip
    ]{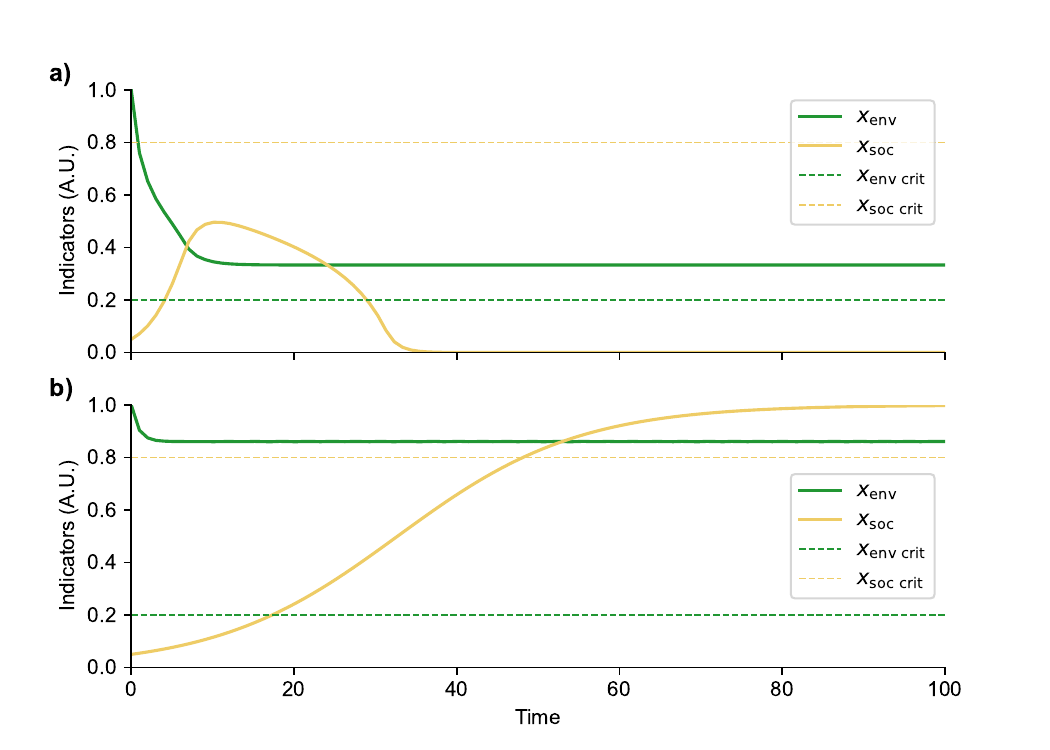}
    \caption{
    {\textbf{Examples of model behaviour from our chosen toy model.} The two plots show the time dyamics of the socio-economic ($x_{\text{soc}}$) and environmental ($x_{\text{env}}$) indicators, with their respective critical thresholds ($x_{\text{soc crit}}$ and $x_{\text{env crit}}$), for two different model input parameter configurations (parameters $c$ and $\eta$). \textbf{a)} While natural resources are not critically compromised, socio-economic targets are not met ($c=0.42$, $\eta=0.9$). \textbf{b)} Both indicators achieve their respective thresholds, i.e. socio-economic targets are met without compromising natural resources ($c=0.2$, $\eta=0.9$). 
    }
    }
    \label{fig:model_dynamics}
\end{figure}

\subsection{\textcolor{black}{Scores from indicators}}\label{supplementary_material_indicators}

We define two performance scores ($N=2$) from our computational model as the average of the environmental and social indicators respectively: 

\begin{itemize}
    \item Environmental performance $v_1 = \frac{1}{T} \int_{0}^{T} x_{\text{env}}\textcolor{black}{-x_{\text{env-crit}}}(t) \, dt$,
    \item Socio-economic performance $v_2 = \frac{1}{T} \int_{0}^{T} x_{\text{soc}}-x_{\text{soc-crit}} \, dt$,
\end{itemize}

where

\begin{itemize}
    \item \textcolor{black}{$x_{\text{env-crit}}$ is a tipping point below which the environmental resource ceases to regenerate.}
    \item $x_{\text{soc-crit}}$ is the minimum social threshold required to meet basic needs (the equivalent of the `social foundation' in the Doughnut).
\end{itemize}

\subsection{Machine Learning}\label{sec:ML_parameters}

\subsubsection{Random Forest Classifier}\label{sec:random_forest_classifier}

We sample synthetic data using \texttt{from emukit.core.initial\_designs import RandomDesign}, with 500 samples, setting \texttt{np.random.seed(42)}. We split test and training data using \texttt{from sklearn.model\_selection import train\_test\_split} with \texttt{random\_state = 42}.
    
For the RFC, we applied the following hyperparameters: max depth = 3; 100 trees; stratified data for training. We set max depth to 3 to pursue frugality and interpretability. Future work may wish to compare the performance of this classification with performance from a hyperparameter-optimised RFC.

Stratified 5-fold cross-validation on the RFC achieves $0.968 \pm 0.013$ accuracy. When sampling the model a high number of times (e.g. 5000), the fraction of samples falling outside the Doughnut is $0.912$, denoting high class imbalance.

\begin{figure}[htp!]
    \centering
    \includegraphics[width=\linewidth, 
    trim={0.0cm, 0.3cm, 0.0cm, 0.2cm}, clip
    ]{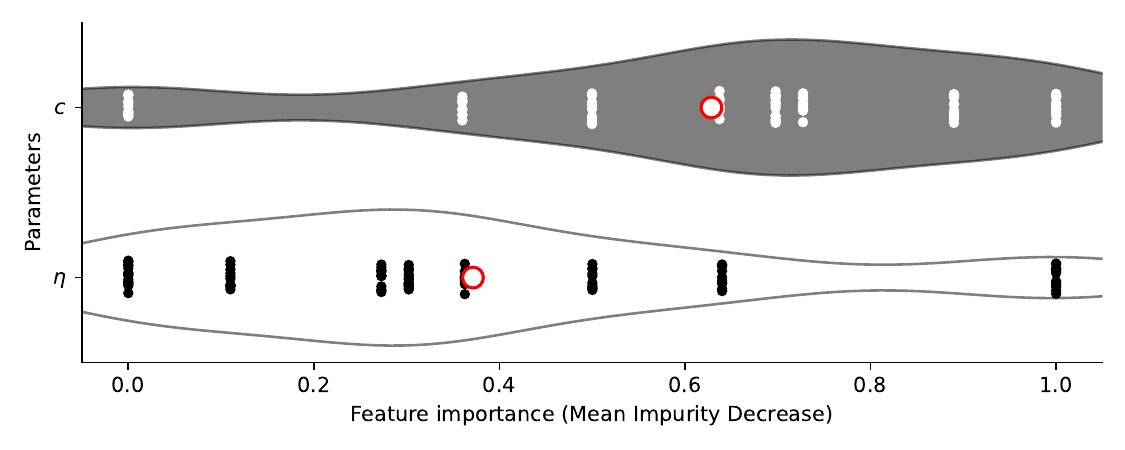}
    \caption{\textbf{Feature importance computed from the RFC.}}
    \label{fig:feature_importance}
\end{figure}

We complement our RFC analysis by adding feature importance ranking (Fig. \ref{fig:feature_importance}). Feature importance values can be obtained for free from RFCs. We find that the consumption parameter $c$ impacts the classification of the model dynamics as `within' or `outside' the Doughnut more than the efficiency parameter $\eta$.

\subsubsection{$Q$-learning}\label{sec:q_learning}
For the RL agent, we applied the following parameters: learning rate $\alpha=0.1$, discount factor $\gamma=0.5$, inverse temperature $\beta=2$ for softmax decision rule, 30,000 training episodes, 50 steps. 

RL states $s$ are defined by discretising the parameter space in 10 values for each model input parameter ($c$ and $\eta$).


At each time step $t$ within an episode, the agent selects an action $a_t$ according to a softmax decision rule:

\begin{equation}
    P(a_t \mid c_t, \eta_t) = \frac{e^{\beta Q(c_t, \eta_t, a_t)}}{\sum_{a' \in \mathcal{A}} e^{\beta Q(c_t, \eta_t, a')}}    
\end{equation}

We define reward as:

\begin{equation}\label{eq:reward_q}
    r_t = R(c_{t+1}, \eta_{t+1}) + \gamma \max_{a' \in \mathcal{A}(c_{t+1}, \eta_{t+1})} Q(c_{t+1}, \eta_{t+1}, a'),
\end{equation}

where

\begin{equation}\label{eq:reward_q_2}
    R(c_{t+1}, \eta_{t+1}) = D(c_{t+1}, \eta_{t+1}),
\end{equation}

with $D$ defined in eq. \ref{eq:doughnut} and computed with the parameter values where the state $s$ is centred.

Finally, the $Q$-value update rule is:
\begin{equation}
    Q(c_{t}, \eta_{t}, a_t) \xleftarrow{} Q(c_t, \eta_t, a_t) + \alpha \big( r_t - Q(c_t, \eta_t, a_t) \big).
\end{equation}
    

\subsection{Agreement score}\label{sec:agreement-computation}


We train a Random Forecast Classifier on a binary classification task, to classify whether a given configuration of model parameter values produces a model regime falling inside or outside the Doughnut (see \nameref{methods}).

We develop an \textit{agreement} score leveraging the trained RFC to: 1) identify the parameter ranges that produce the desired model regimes, 2) rank them, and 3) present them in readily interpretable tabular form. The algorithm works by first identifying suitable parameter ranges, to then compute agreement for each range.

\subsubsection{Identify split thresholds}
In a trained RFC, each decision tree defines its own split thresholds for features. We collect all thresholds across trees and count their frequency.



\subsubsection{Select split thresholds}

\begin{figure}[htp!]
    \centering
    \includegraphics[width=\linewidth, 
    trim={0.0cm, 0.0cm, 0.0cm, 0.0cm}, clip
    ]{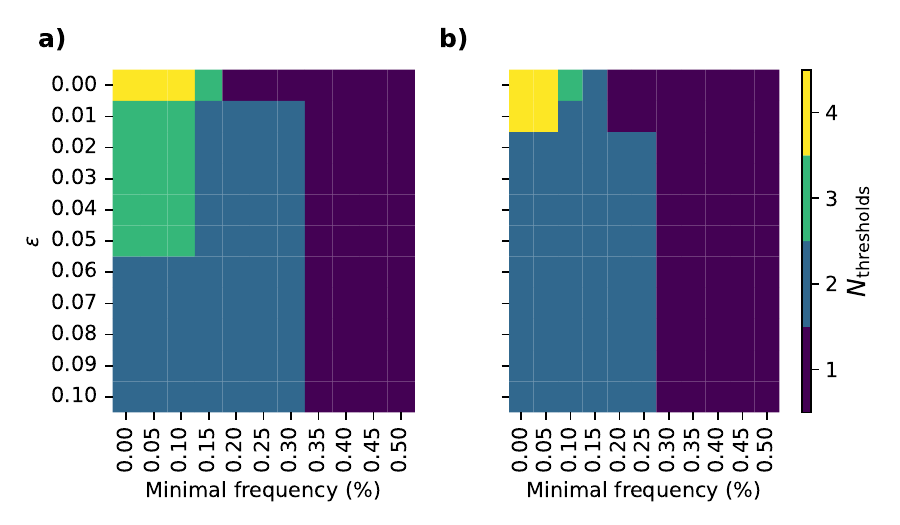}
    \caption{
    \textbf{Heatmap showing how the number of selected split-thresholds, collected among trees of a Random Forest Classifier trained on our task, changes according to the parameter $\epsilon$  and the minimal frequency.} Each plot refers to a different input model parameter: $c$ in \textbf{a} and $\eta$ in \textbf{b}. The higher the number of selected thresholds per parameter, the more granular, but also less interpretable.
    }
    \label{fig:split_threshold}
\end{figure}

For each feature, we sort the split thresholds by frequency. Within each feature $i$, starting from the most frequent split threshold, we merge all thresholds lying within a chosen $\varepsilon_i$ ($\varepsilon=0.02$ for all features, in our case), keeping the split threshold with the highest count. Merged split thresholds also accumulate their counts. We then set a percentage ($25\%$ in our case) as the minimal frequency threshold to retain split thresholds. All split threshold with a lower frequency are discarded. The remaining split thresholds define the selected features' ranges (in our case, model parameter ranges), which form multi-feature bins. Here, the selection step, which might seem optional, not only helps to deal with a more reasonable number of feature splits, but also plays a central role in creating large-enough value ranges to bin test data and compute test accuracy in later steps of the algorithm, especially for high-dimensional input spaces (Fig. \ref{fig:split_threshold}). This selection step can be further optimised to ensure that all bins contain a minimum number of test data points.

\subsubsection{Compute agreement within feature ranges}

We sample a high number (e.g. 100,000) of model input parameter configurations as input feature values. We then bin them into the bins obtained in the previous step. For each bin $b$ and for each tree $\tau$ we collect the predicted class of each sample within that bin. Next, for each tree, we average the predicted classes $\hat{c}_n \in \{0,1\}$ (with 1 indicating that the input features produce a Doughnut-compliant model output, 0 otherwise) across $N_b$ samples within that bin. We define this measure as `estimated Doughnut frequency' $\hat{f}_{\text{raw}, \tau, b}$ for that bin:
\begin{equation}
    \hat{f}_{\text{raw}, \tau, b} = \frac{1}{N_b} \sum_{n=1}^{N_b}{\hat{c}_n}.
\end{equation}
We then compute the raw accuracy $a_{\text{raw}, \tau, b}$ for each tree, for all labelled test data within a given bin.

Since we are dealing with a classification task with only two classes, we can leverage both high and low accuracy. The further below 50\% the test accuracy is, the stronger our confidence that the actual class is the one \textit{not} predicted by the tree. We can therefore compute a `useful' accuracy

\begin{equation}
    a_{\text{useful}, \tau, b} = 2\, |a_{\text{raw}, \tau, b} - 0.5|,
\end{equation}

and a `useful' estimated Doughnut frequency

\begin{equation}
    \hat{f}_{\text{useful}, \tau, b} = 
        \begin{cases}
        \hat{f}_{\text{raw}, \tau, b}, & a_{\text{raw}, \tau, b} > 0.5 \\
        1 - \hat{f}_{\text{raw}, \tau, b}, & a_{\text{raw}, \tau, b} < 0.5
        \end{cases}    
\end{equation}

for each tree $\tau$ and bin $b$.

Next, for each bin, we want to emphasise the frequency values of the Doughnut $\hat{f}_{\text{raw}, \tau, b}$ from the trees with the highest useful accuracy. We therefore normalise useful accuracies across trees, bin-wise, using softmax:

\begin{equation}
    a_{\text{norm}, \tau, b} = \frac{e^{\beta_{\text{norm}} \, a_{\text{useful}, \tau, b}}}{\sum_{\tau'} e^{\beta_{\text{norm}} \, a_{\text{useful}, \tau', b}}},    
\end{equation}

setting the inverse temperature parameter $\beta_{\text{norm}}=1$.



We then measure agreement $A$ for each bin by averaging the estimated Doughnut frequency across trees, weighted by their own normalised useful accuracy:

\begin{equation}\label{eq:agreement}
    \text{A}_b = \sum_{\tau} a_{\text{norm},\tau, b} \cdot \hat{f}_{\text{useful}, \tau, b}.
\end{equation}

Finally, agreement is scaled to lie between $[-1, 1]$, parameter ranges are sorted, and presented in tabular form (Fig. \ref{fig:combined-panel}a).


\subsection{Hardware and computation}\label{sec:Hardware}

We used a Dell Precision 7760 laptop with a 2.60 GHz 16-core Intel Xeon(R) W-11955M processor. Generating the 500 samples (i.e. running the model 500 times) and training the classifier takes $<1.4$ seconds. Running the agreement score algorithm (section \ref{sec:agreement-computation}) adds less than 2.5 seconds. Training the $Q$-learning agent required less than 1 min 30 seconds.\\

The full research project involved trial-and-error to adapt and optimise the ML tools to the specific computational experiments. We only report the final version of our experiments.

\subsection{Code}
We implemented our statistical pipeline in Python 3.12.3. The code is available at \href{https://github.com/stefanovrizzi/doughnut_ml/tree/master}{https://github.com/stefanovrizzi/doughnut\_ml}.

\end{document}